\pdfoutput=1

\documentclass[11pt]{article}

\usepackage{EMNLP2022}

\usepackage{times}
\usepackage{latexsym}

\usepackage[T1]{fontenc}

\usepackage[utf8]{inputenc}

\usepackage{microtype}

\usepackage{inconsolata}

\usepackage{amsmath}
\usepackage{cleveref}

\usepackage{array}
\usepackage{makecell}
\usepackage{color}
\usepackage{soul}

\newcommand{\n}{$\backslash$n}

\title{UniHD at TSAR-2022 Shared Task:\\ Is Compute All We Need for Lexical Simplification?}

\author{Dennis Aumiller \and Michael Gertz \\
  Institute of Computer Science \\
  Heidelberg University \\
  \texttt{\{aumiller, gertz\}@informatik.uni-heidelberg.de}}

\begin{document}
\maketitle
\begin{abstract}
Previous state-of-the-art models for lexical simplification consist of complex pipelines with several components, each of which requires deep technical knowledge and fine-tuned interaction to achieve its full potential.
As an alternative, we describe a frustratingly simple pipeline based on prompted GPT-3 responses, beating competing approaches by a wide margin in settings with few training instances.
Our best-performing submission to the English language track of the TSAR-2022 shared task consists of an ``ensemble'' of six different prompt templates with varying context levels. As a late-breaking result, we further detail a language transfer technique that allows simplification in languages other than English. Applied to the Spanish and Portuguese subset, we achieve state-of-the-art results with only minor modification to the original prompts.
Aside from detailing the implementation and setup, we spend the remainder of this work discussing the particularities of prompting and implications for future work.
Code for the experiments is available online.\footnote{\url{https://github.com/dennlinger/TSAR-2022-Shared-Task}}
\end{abstract}

\section{Introduction}
With recent advancements in Machine Learning (ML) research coming largely from increasing compute budgets, Richard Sutton coined the idea of a ``bitter lesson'', wherein more computational power will ultimately supersede a hand-crafted solution~\cite{sutton-2019-bitter}.
More recently, increasing compute power on a general purpose architecture has also shown to be wildly successful in the Natural Language Processing (NLP) community~\cite{vaswani-etal-2017-attention,wei-etal-2022-emergent}.
In particular, emergent capabilities in very large language models (vLLMs) have made it possible to approach a variety of tasks wherein only few (if any) samples are labeled, and no further fine-tuning on task-specific data is required at all.\\
In stark contrast to the complex pipelines in modern lexical simplification systems~\cite{ferres-etal-2017-adaptable,qiang-etal-2020-lexical,stajner-etal-2022-lexical}, we present a simplistic approach utilizing few-shot prompts based on a vLLM with basic instructions on simplification, which returns frustratingly good results considering the overall complexity of the approach, which utilizes a grand total of four hand-labeled instances.
We present our results on the TSAR-2022 shared task~\cite{tsar-2022-findings}, which evaluates lexical simplification systems in three available languages (English, Spanish and Portuguese), with ten labeled instances and around 350 unlabeled test samples provided per language.
For the English subset, official results rank our model as the best-performing submission, indicating that this approach may be another instance of the bitter lesson.\\
While the initial findings are indeed promising, we want to carefully evaluate erroneous instances on the test set to analyze potential pitfalls, and further detail some of our experiences in hand-crafting prompts.
We also acknowledge the technical challenges in reproducing (and deploying) systems based on vLLMs, especially given that suitable models exceed traditional computing budgets.


\section{Prompt-based Lexical Simplification}
With the public release of the GPT-3 language model~\cite{brown-etal-2020-language}, OpenAI has started the run on a series of now-available vLLMs for general-purpose text generation~ \cite{thoppilan-etal-2022-lambda,bigscience-2022-bloom,zhang-etal-2022-opt}. Across these models, a general trend in scaling beyond a particular parameter size can be observed, while keeping the underlying architectural design close to existing smaller models.
Through exhibiting zero-shot transfer capabilities, such models have also become more attractive for lower-resourced tasks; oftentimes, models are able to answer questions formulated in natural language with somewhat sensible results. Particular template patterns (so-called \emph{prompts}) are frequently used to guide models towards predicting a particularly desirable output or answer format, without requiring a dedicated training on labeled examples.

\noindent Utilizing this paradigm shift, we experimented with different prompts issued to OpenAI's largest available model, \texttt{text-davinici-002}, which totals 176B parameters.
Our first approach uses a singular prompt template in a zero-shot setting to obtain predictions for the shared task; we further improve upon these results by combining predictions from different prompt templates later on.

\subsection{Run 1: Zero-shot Prediction}
Upon inspecting the provided trial data, we noted that the simplification operations required a vastly different contextualization within the provided sample sentence. Whereas some instances can be solved with pure synonym look-ups (e.g., ``compulsory'' and ``mandatory''), others require a more nuanced look at the context sentence (e.g., replacing ``disguised'' with ``dressed'').
To avoid biasing system predictions by providing samples as a prompt template, we provide a baseline that is entirely based on a single zero-shot query; it provides the context sentence and identifies the complex word, asking the model for ten simplified synonyms of the complex word in the given context.
Given that no additional knowledge is provided to the model, the zero-shot contextual query also provides a reasonable lower-bound for the task setting.
A secondary advantage of minimal provided context in zero-shot settings is the reduced computational cost, which will be discussed in more detail in 
\Cref{sec:budget}.

\subsection{Filtering Predictions}
Model suggestions are returned as free-form text predictions, generally in the form of comma-separated lists or enumerations.
This requires the additional step of parsing the output prediction into the more structured ranked predictions required for the shared task, which varies between the models used.
In our experience, no clear pattern can be expected from the model and seems to be non-deterministic even with set template structures.
We additionally employ a list of simple filters to ensure the quality of predictions, as detailed in \Cref{sec:filters}.
The resulting model suggestions are considered in ranked order, and no prediction confidence scores or similar information was used to re-rank single-prompt predictions.

\subsection{Run 2: Ensemble Predictions}
Upon inspecting the results from the first run, we noticed that in some instances, predictions were almost fully discarded due to filtering.
Simultaneously, we had already previously encountered strong variability in system generations when changing the prompt template or altering the context setting.
To this extent, an ensemble of predictions from multiple different prompt templates was utilized to broaden the spectrum of possible generations, as well as ensuring that a minimum number of suggestions survives the filtering step.

\subsubsection{Prompt Variations}

The exact prompts are detailed in \Cref{tab:prompts}.
Utilized variations can be grouped into \emph{with context} (the context sentence is provided), or \emph{without context} (synonyms are generated from the complex word alone).
Simultaneously, different prompts also contain between zero and two examples taken from the trial data, including their expected outputs. This can be interpreted as a few-shot setting in which the model is demonstrated on what correct answers may look like for the particular task.
We further vary the generation temperature, where a higher value increases the likelihood of a more creative (but not always correct) prediction, enabling a more diverse candidate set.

\subsubsection{Combining Predictions}

For each of the six prompts $p$, we ask the model to suggest ten alternative simplified expressions $S_p$ and filter them with the exact same rules as the single prompt system in Run 1. In order to combine and re-rank suggestions $s$, we assign a combination score $V$ to each distinct prediction $s \in \bigcup_p S_p$:
\begin{align}
	V(s) = \sum_p \text{max}(5.5 - 0.5 \cdot \text{rank}_{S_p}(s), 0),
\end{align}
where $\text{rank}_{S_p}(s)$ is the ranked position of suggestion $s$ in the resulting ranking from prompt $p$. If $s \notin S_p$, we set $\text{rank}_{S_p}(s) = \infty$. The scaling parameters are chosen arbitrarily and can be adjusted to account for the expected number of suggestions per prompt.
We estimate that the biggest performance improvement is coming simply from providing enough predictions post filtering.
As a secondary gain, we see more consistent behavior in the top-most prediction slots, boosting especially the @1 performance of the ensemble.

\begin{table*}
	\hspace*{-0.25cm}
	\setlength{\tabcolsep}{3pt}
\centering
\begin{tabular}{l|c|ccc|ccc|ccc}
 	&       & \multicolumn{3}{c|}{\textbf{Acc@k@Top1}} & \multicolumn{3}{c|}{\textbf{MAP@k}} & \multicolumn{3}{c}{\textbf{Potential@k}}\\
\textbf{Run} & \textbf{ACC@1} & $k=1$ & $k=2$ & $k=3$ & $k=3$ & $k=5$ & $k=10$ & $k=3$ & $k=5$ & $k=10$\\
\hline
Ensemble (Ours) & \textbf{0.8096} & \textbf{0.4289} & \textbf{0.6112} & \textbf{0.6863} & \textbf{0.5834} & \textbf{0.4491} & \textbf{0.2812} & \textbf{0.9624} & \textbf{0.9812} & \textbf{0.9946}\\
Single (Ours) & 0.7721 & 0.4262 & 0.5335 & 0.5710 & 0.5090 & 0.3653 & 0.2092 & 0.8900 & 0.9302 & 0.9436\\
\hline
MANTIS-1 & 0.6568 & 0.319 & 0.4504 & 0.5388 & 0.473 & 0.3599 & 0.2193 & 0.8766 & 0.9463 & 0.9785\\
UoM\&MMU-1 & 0.6353 & 0.2895 & 0.4530 & 0.5308 & 0.4244 & 0.3173 & 0.1951 & 0.8739 & 0.9115 & 0.9490\\
LSBert & 0.5978 & 0.3029 & 0.4450 & 0.5308 & 0.4079 & 0.2957 & 0.1755 & 0.8230 & 0.8766 & 0.9463 \\
TUNER & 0.3404 & 0.1420 & 0.1689 & 0.1823 & 0.1706 & 0.1087 & 0.0546 & 0.4343 & 0.4450 & 0.4450 \\
\end{tabular}
\caption{Results on the English language test set of the TSAR-2022 shared task, ranked by \emph{ACC@1} scores. Listed are our own results (\emph{Ensemble} and  \emph{Single}), the two best-performing competing systems~(\emph{MANTIS} and \emph{UoM\&MMU}), as well as provided baselines (\emph{LSBert}~\cite{qiang-etal-2020-lexical} and TUNER~\cite{ferres-etal-2017-adaptable}).}
\label{tab:english}
\end{table*}

\section{Results and Limitations}

\subsection{Results for English}
For the official runs, we initially only submitted predictions for the English subset; an excerpt of the results can be seen in \Cref{tab:english}.
While the zero-shot single prompt run has consistently better results on most metrics, it does not outperform all systems for large candidate sets; e.g., Potential@10 is lower than that of competing approaches, including the LSBert baseline.
We attribute this to the previously mentioned issue of filtering predictions, and can see a consequent improvement especially for larger $k$ by using the proposed ensemble method. Here, the Potential@10 scores indicate that at least one viable prediction is present in \emph{all but three samples}.

\subsection{Results for Spanish and Portuguese}
Given the surprisingly good results on the English subset, we decided to extend our experiments to the Spanish and Portuguese tracks as well.
Transferring the prompts to Spanish or Portuguese is surprisingly simple. We alter the prompt to:
\emph{``Given the above context, list ten alternative \textbf{Spanish} words for `\texttt{complex\_word}' that are easier to understand.''} (bold highlight indicates change).\\
Without this adaption, returned suggestions generally tend to be in English, which could be an attractive opportunity to mine cross-lingual simplifications in future work. By adding the output language explicitly, we ensure that the suggestions match the expected results.
For Portuguese, the prompt can be adapted accordingly.

\noindent We find that our system also outperforms all competing submitted approaches in the shared task; result comparisons can be found in \Cref{tab:spanish} and \ref{tab:portuguese} in the Appendix, respectively.
Notably, predictions for Portuguese perform slightly better, which goes against intuition, given that Spanish is usually a highly represented language in multilingual corpora.
We suspect that a more literal wording of synonyms in Portuguese, compared to multi-word expressions in Spanish, could be the cause.

\subsection{Error Analysis}

\begin{table*}[ht]
	\hspace*{-0.35cm}
	\setlength{\tabcolsep}{4pt}
	\centering
	\begin{tabular}{l|l|l}
		\textbf{Error Type} & \textbf{Context (complex word in bold)} & \textbf{Model Predictions} \\
		\hline
		\textbf{Lack of Context} & \makecell[l]{\#7-8 Despite the fog, other flights are reported\\to have landed safely leading up to the \textbf{collision}.} & car crash, train wreck, ... \\
		\hline
		\textbf{Hallucinations} & \makecell[l]{The larva grows to about 120-130 mm,\\and \textbf{pupates} in an underground chamber.} & Transforms into a pupa, ... \\
		\hline
		\textbf{Language} & \makecell[l]{[...] propiciado la \textbf{decadencia} de la Revolución francesa.} & decline, deterioration, ...
	\end{tabular}
	\caption{Instances of observed failure classes in our system's predictions.}
	\label{tab:failures}
\end{table*}

As is common for sequence-to-sequence tasks, crafting an approach centered around a LM requires consideration of the particular challenges arising.
We detail some of the errors we have encountered in our predictions that are unlikely to appear in more stringently designed pipelines.
Instances for particular failure cases can be found in \Cref{tab:failures}.

\paragraph{Unstable Prompts}
One of the primary challenges, particularly for zero-shot prompt settings, is the unreasonable variance observed in results based on even just slightly altered prompt templates.
For example, when removing the explicit mention of \emph{Context:}, \emph{Question:} and \emph{Answer:} in the prompt template, the model is frequently predicting fewer than the ten requested answers.
Practical limitations in our computational budget also mean that we have no guarantee that these prompts are yielding the best possible results; given the variability, multiple runs should be compared for a thorough pattern of a ``best'' prompt.

\paragraph{Lack of Context}
Instances with longer (or subtly enforced) context cues show issues where these hints are not properly recognized. In \Cref{tab:failures}, we can see the model changing the term ``\emph{collision}'' to a particular mode of transportation, such as ``\emph{car crash}'', while an explicit context clue is given through the word ``\emph{flight}'' in the original sentence.

\paragraph{Enforcing Language}
While the transfer to Spanish and Portuguese is largely successful, the model's capabilities seem to be still limited in maintaining the language throughout all samples. For instances with particularly rare complex terms, the predictions are sometimes still in English, despite the specific prompt request to return Spanish/Portuguese results.

\paragraph{Hallucinations}
The necessity for post-filtering of suggestions stems largely from the spontaneous occurrence of hallucinations in responses. While hallucinations in vLLMs are less about invalid vocabulary terms, we observe instances where unnecessary multi-word suggestions were chosen over a simple synonymous single-word expression, or random inflections (such as the infinitive form with an additional ``to'') were generated.\\
Similar to the issues with language enforcing, this occurs more frequently with particularly complex words; in this sense, the system conversely fails at instances that are most in need of simplification.
However, we note that some of the generated multi-word expressions are actually more helpful for understanding, even though the generations are not precisely matching expected outputs.


\subsection{Computational Limitations}
\label{sec:budget}
Running a vLLM in practice, even for inference-only settings, is non-trivial and requires compute resources that are far beyond many public institution's hardware budget. For the largest models with publicly available checkpoints\footnote{At the time of writing, this would be the 176B Bloom model~\cite{bigscience-2022-bloom}, which has a similar parameter count to OpenAI's davinci-002 model.}, a total of around \textbf{325GB GPU memory} is required, assuming efficient storage in \texttt{bfloat16} or similar precision levels.
The common alternative is to obtain predictions through a (generally paid) API, as was the case in this work. Especially for the ensemble model, which issues six individual requests to the API per sample, this can further bloat the net cost of a single prediction.
To give context of the total cost, we incurred a total charge of slightly over \$7 for computing predictions across the entire test set of 373 English samples, which comes out to about 1000 tokens per sample, or around \$0.02 at the current OpenAI pricing scheme.\footnote{\url{https://openai.com/api/pricing/}, last accessed: 2022-10-01}
For the Spanish subset and language-dependent prompt development, the total cost came to about \$10, primarily due to longer sample contexts. Costs for Portuguese processing were around \$6.50.
While the singular prompt approach is cheaper at around 1/6 of the total cost, even then a continuously deployed model has to be supplied with a large enough budget.
Aside from monetary concerns, environmental impacts are also to be considered for larger-scale deployments of this kind~\cite{lacoste-etal-2019-quantifying}.

\section{Conclusion and Future Work}
Utilizing prompted responses from vLLMs seems to be a promising direction for lexical simplification; particularly in the constrained setting with pre-identified complex words the model performs exceptionally well, even when presented with a severely restricted budget of labeled training data.
While the approach also offers promising directions for multi- and cross-lingual approaches, obtaining state-of-the-art results in other languages, we are presented with a prohibitive amount of computation per sample instance.
It would therefore be an interesting addition to deal with resource-constraint systems, putting the prediction power into a slightly different perspective.
Finally, we are reminded of the unstable nature of neural LMs; given similar inputs, quality can vary greatly between samples, including a complete breakdown in performance.\\
For future work, we are considering approaches to generate static resources from vLLMs~\cite{schick-schutze-2021-generating}, which may require only a one-time commitment to spending on datasets, which can then used as training data for cheaper systems.
Exploration of prompt tuning approaches for automated search of suitable prompt templates would also greatly accelerate the development process of domain-specific applications~\cite{lester-etal-2021-power}.

\bibliography{anthology,custom}
\bibliographystyle{acl_natbib}

\appendix
\newpage

\section{Prompt Templates}
\label{sec:prompts}

\Cref{tab:prompts} provides the exact prompt templates used in the submission. Notably, the \emph{zero-shot with context} prompt is included twice, but with different generation temperatures; with this we increase the likelihood of strong candidates being retained.
For few-shot prompts, we have taken samples from the previously published trial set for the respective language. In instances where less than 10 distinctly different suggestions were provided by annotators, we manually extended the list of examples to match exactly ten results based on our own judgment. For instances with more provided suggestions, we limit ourselves to the ten most frequently occurring ones. The reason for this is that GPT-3 otherwise tended to return an inconsistent number of suggestions in our preliminary testing. The exact prompts for the Spanish and Portuguese runs can be found in our repository.

\begin{table*}
	\sloppy
	\hspace*{-0.95em}
	\setlength{\tabcolsep}{2pt}
	\begin{tabular}{|l|l|}
		\hline
		\textbf{Prompt Type} & \textbf{Template} \\
		\hline
		\textbf{Zero-shot /w context} & \texttt{Context: \{context\_sentence\}\n} \\
		temperature: 					   & \texttt{Question: Given the above context, list ten alternatives for} \\
		0.3 (conservative),					   & \texttt{    ``\{complex\_word\}'' that are easier to understand.\n} \\
		0.8 (creative)					   &\texttt{Answer:} \\
							   
		\hline
		\hline
		\textbf{Single-shot /w context} & \texttt{Context: A local witness said a separate group of attackers}\\
		temperature: 0.5		& \texttt{ disguisedin burqas — the head-to-toe robes worn by conservative}\\
								& \texttt{ Afghan women — then tried to storm the compound.\n}\\
								& \texttt{Question: Given the above context, list ten alternative words }\\
								& \texttt{ for ``disguised'' that are easier to understand.\n}\\
								& \texttt{Answer:\n1. concealed\n2. dressed\n3. hidden\n4. camouflaged\n}\\
								& \texttt{    5. changed\n6. covered\n7. masked\n8. unrecognizable\n}\\
								& \texttt{ 9. converted\n10. impersonated\n\n}\\
								& \texttt{Context: \{context\_sentence\}\n} \\
								& \texttt{Question: Given the above context, list ten alternatives for} \\
								& \texttt{    ``\{complex\_word\}'' that are easier to understand.\n} \\
								&\texttt{Answer:} \\
		\hline
		\textbf{Two-shot /w context}  & \texttt{Context: That prompted the military to deploy its largest }\\
		temperature: 0.5& \texttt{warship, the BRP Gregorio del Pilar, which was recently }\\
		& \texttt{acquired from the United States.\n}\\
		& \texttt{Question: Given the above context, list ten alternative words }\\
		& \texttt{ for ``deploy'' that are easier to understand.\n}\\
		& \texttt{Answer:\n1. send\n2. post\n3. use\n4. position\n5. send out\n}\\
		& \texttt{ 6. employ\n7. extend\n8. launch\n9. let loose\n10. organize\n\n}\\
		& \texttt{Context: The daily death toll in Syria has declined as the}\\
		& \texttt{ number of observers has risen, but few experts expect the }\\
		& \texttt{ U.N. plan to succeed in its entirety.\n}\\
		& \texttt{Question: Given the above context, list ten alternative words }\\
		& \texttt{ for ``observers'' that are easier to understand.\n}\\
		& \texttt{Answer:\n1. watchers\n2. spectators\n3. audience\n4. viewers\n}\\
		& \texttt{ 5. witnesses\n6. patrons\n7. followers\n8. detectives\n}\\
		& \texttt{ 9. reporters\n10. onlookers\n\n}\\
		& \texttt{Context: \{context\_sentence\}\n} \\
		& \texttt{Question: Given the above context, list ten alternatives for} \\
		& \texttt{    ``\{complex\_word\}'' that are easier to understand.\n} \\
		&\texttt{Answer:} \\
		\hline
		\textbf{Zero-shot w/o context} & \texttt{Give me ten simplified synonyms for the following word:} \\
		temperature: 0.7 & \texttt{\{complex\_word\}} \\
		\hline
		\textbf{Single-shot w/o context} & \texttt{Question: Find ten easier words for ``compulsory''.\n} \\
		temperature: 0.6						    & \texttt{Answer:\n1. mandatory\n2. required\n3. essential\n4. forced\n}\\
								    & \texttt{    5. important\n6. necessary\n7. obligatory\n8. unavoidable\n} \\
								    & \texttt{    9. binding\n10. prescribed\n\n} \\
								    & \texttt{Question: Find ten easier words for ``\{complex\_word\}''.\n}\\
								    & \texttt{Answer:}\\
		\hline
		
	\end{tabular}
	\caption{The English prompt templates used for querying the OpenAI model, including associated generation temperatures. Only written out ``\texttt{\n}'' symbols indicate newlines, visible line breaks are inserted for better legibility. Only top-most prompt template with conservative temperature was used in the single prompt (Run 1), as well as in the ensemble run (Run 2). All other prompts were only included in the ensemble submission.}
	\label{tab:prompts}
\end{table*}

\section{Hyperparameters}
We use the OpenAI Python package\footnote{\url{https://github.com/openai/openai-python}} version 0.23.0 for our experiments. For generation, the function \texttt{openai.Completion.create()} is used, where most hyperparameters remain fixed across all prompts. We explicitly list those hyperparameters below that differ from their respective default values.
\begin{enumerate}
	\item \texttt{model="text-davinci-002"}, which is the latest and biggest available model for text completion.
	\item \texttt{max\_tokens=256}, to ensure sufficient room for generated outputs. In practice, most completions are vastly below the limit.
	\item \texttt{frequency\_penalty=0.5}, as well as \texttt{presence\_penalty=0.3}, which jointly penalize present tokens and token repetitions. The values are well below the maximum (values can range from -2 to 2), since individual subword tokens might indeed be present several times across multiple (valid) predictions. A more detailed computation can be found in the documentation of OpenAI.\footnote{\url{https://beta.openai.com/docs/api-reference/parameter-details}}
\end{enumerate}

\noindent Outside of the repetition penalties, the most influential parameter choice for generation is the sampling temperature. We generally take a more measured approach than the default (\texttt{temperature=1.0}), but vary temperature across our ensemble prompts to ensure a more diverse result set overall. We list the used temperatures in \Cref{tab:prompts}. \emph{Zero-shot with context} is used twice in the ensemble, once with a more conservative temperature, and once with a more ``creative'' (higher) temperature. For the singular prompt run, we use the conservative \emph{zero-shot with context} variant.

\section{Post-Filtering Operations}
\label{sec:filters}
Given the uncertain nature of predictions by a language model, we employ a series of post-filtering steps to ensure high quality outputs. This includes stripping newlines/spaces/punctuation (\texttt{$\backslash$n :;.?!}), lower-casing, removing infinitive forms (in some instances, we observed predictions in the form of ``to deploy'' instead of simply ``deploy''), as well as removing identity predictions (e.g., the prediction being the same as the original complex word) and deduplicating suggestions. Additionally, we noticed that for some instances, generated synonyms resemble more of a ``description'' rather than truly synonymous expressions (example: ``people that are crazy'' as a suggestion for ``maniacs''). Given the nature of provided data, we removed extreme multi-word expressions (for English, any suggestion with more than two words, for Spanish and Portuguese more than three words in a single expression).

\begin{table*}
	\hspace*{-0.35cm}
	\setlength{\tabcolsep}{3pt}
	\centering
	\begin{tabular}{l|c|ccc|ccc|ccc}
		&       & \multicolumn{3}{c|}{\textbf{Acc@k@Top1}} & \multicolumn{3}{c|}{\textbf{MAP@k}} & \multicolumn{3}{c}{\textbf{Potential@k}}\\
		\textbf{Run} & \textbf{ACC@1} & $k=1$ & $k=2$ & $k=3$ & $k=3$ & $k=5$ & $k=10$ & $k=3$ & $k=5$ & $k=10$\\
		\hline
		Ensemble (Ours) & \textbf{0.6521} & \textbf{0.3505} & \textbf{0.5108} & \textbf{0.5788} & \textbf{0.4281} & \textbf{0.3239} & \textbf{0.1967} & \textbf{0.8206} & \textbf{0.8885} & \textbf{0.9402}\\
		Single (Ours) & 0.5706 & 0.3070 & 0.3967 & 0.4510 & 0.3526 & 0.2449 & 0.1376 & 0.6902 & 0.7146 & 0.7445\\
		\hline
		PresiUniv-1 & 0.3695 & 0.2038 & 0.2771 & 0.3288 & 0.2145 & 0.1499 & 0.0832 & 0.5842 & 0.6467 & 0.7255\\
		UoM\&MMU-3 & 0.3668 & 0.1603 & 0.2282 & 0.269 & 0.2128 & 0.1506 & 0.0899 & 0.5326 & 0.6005 & 0.6929\\
		LSBert & 0.2880 & 0.0951 & 0.1440 & 0.1820 & 0.1868 & 0.1346 & 0.0795 & 0.4945 & 0.6114 & 0.7472\\
		TUNER & 0.1195 & 0.0625 & 0.0788 & 0.0842 & 0.0575 & 0.0356 & 0.0184 & 0.144 & 0.1467 & 0.1494 \\
	\end{tabular}
	\caption{Results on the Spanish language test set of the TSAR-2022 shared task, ranked by \emph{ACC@1} scores. Listed are our own results (\emph{Ensemble} and  \emph{Single}), the two best-performing competing systems~(\emph{PresiUniv} and \emph{UoM\&MMU}), as well as provided baselines (\emph{LSBert}~\cite{qiang-etal-2020-lexical} and TUNER~\cite{ferres-etal-2017-adaptable}).}
	\label{tab:spanish}
\end{table*}

\begin{table*}
	\hspace*{-0.35cm}
	\setlength{\tabcolsep}{3pt}
	\centering
	\begin{tabular}{l|c|ccc|ccc|ccc}
		&       & \multicolumn{3}{c|}{\textbf{Acc@k@Top1}} & \multicolumn{3}{c|}{\textbf{MAP@k}} & \multicolumn{3}{c}{\textbf{Potential@k}}\\
		\textbf{Run} & \textbf{ACC@1} & $k=1$ & $k=2$ & $k=3$ & $k=3$ & $k=5$ & $k=10$ & $k=3$ & $k=5$ & $k=10$\\
		\hline
		Ensemble (Ours) & \textbf{0.7700} & \textbf{0.4358} & \textbf{0.5347} & \textbf{0.6229} & \textbf{0.5014} & \textbf{0.3620} & \textbf{0.2167} & \textbf{0.9171} & \textbf{0.9491} & \textbf{0.9786}\\
		Single (Ours) & 0.6363 & 0.3716 & 0.4625 & 0.5160 & 0.4105 & 0.2889 & 0.1615 & 0.7860 & 0.8181 & 0.8422\\
		\hline
		GMU-WLV-1 & 0.4812 & 0.2540 & 0.3716 & 0.3957 & 0.2816 & 0.1966 & 0.1153 & 0.6871 & 0.7566 & 0.8395\\
		Cental-1 & 0.3689 & 0.1737 & 0.2433 & 0.2673 & 0.1983 & 0.1344 & 0.0766 & 0.524 & 0.5641 & 0.6096\\
		LSBert & 0.3262 & 0.1577 & 0.2326 & 0.286 & 0.1904 & 0.1313 & 0.0775 & 0.4946 & 0.5802 & 0.6737\\
		TUNER & 0.2219 & 0.1336 & 0.1604 & 0.1604 & 0.1005 & 0.0623 & 0.0311 & 0.2673 & 0.2673 & 0.2673
	\end{tabular}
	\caption{Results on the Portuguese language test set of the TSAR-2022 shared task, ranked by \emph{ACC@1} scores. Listed are our own results (\emph{Ensemble} and  \emph{Single}), the two best-performing competing systems~(\emph{GMU-WLV} and \emph{Cental}), as well as provided baselines (\emph{LSBert}~\cite{qiang-etal-2020-lexical} and TUNER~\cite{ferres-etal-2017-adaptable}).}
	\label{tab:portuguese}
\end{table*}

\end{document}